%% file: paper.tex
\documentclass[letterpaper, 10 pt, conference]{ieeeconf}  

\IEEEoverridecommandlockouts                              

\usepackage{amsmath} 
\usepackage{amssymb}  

\usepackage{graphicx}
\usepackage{capt-of} 
\usepackage{overpic} 
\usepackage[export]{adjustbox} 
\usepackage{minibox} 
\usepackage{textcomp} 
\usepackage{tabularx}
\usepackage{multirow}
\usepackage{booktabs}
\usepackage{arydshln}
\usepackage{adjustbox} 
\usepackage{pgfplots} 
\usepackage{siunitx}

\makeatletter
\let\NAT@parse\undefined
\makeatother
\usepackage[hyperfootnotes=false]{hyperref}
\usepackage[caption=false, font=scriptsize]{subfig} 

\newcommand\copyrighttext{\footnotesize \textcopyright~2023 IEEE. Personal use of this material is permitted.  Permission from IEEE must be obtained for all other uses, in any current or future media, including reprinting/republishing this material for advertising or promotional purposes, creating new collective works, for resale or redistribution to servers or lists, or reuse of any copyrighted component of this work in other works.
}%

\newcommand\copyrightnotice{%
	\begin{tikzpicture}[remember picture,overlay]
	\node[anchor=south,xshift=0pt,yshift=14pt] at (current page.south) {\fbox{\parbox{\dimexpr\textwidth-\fboxsep-\fboxrule\relax}{\copyrighttext}}};
	\end{tikzpicture}%
}

\title{\LARGE \bf
Exploring Navigation Maps for Learning-Based Motion Prediction
}

\author{Julian Schmidt$^{1, 2}$, Julian Jordan$^{1}$, Franz Gritschneder$^{1}$, Thomas Monninger$^{3}$ and Klaus Dietmayer$^{2}$
\thanks{The research leading to these results is funded by the BMWK within the project "KI Delta Learning" (F\"orderkennzeichen 19A19013A).}
\thanks{$^{1}$Mercedes-Benz AG, Research \& Development, Stuttgart, Germany \newline {\tt\small julian.sj.schmidt@mercedes-benz.com }}%
\thanks{$^{2}$Ulm University, Institute of Measurement, Control and Microtechnology, Ulm, Germany}%
\thanks{$^{3}$Mercedes-Benz Research \& Development North America, Sunnyvale, CA, USA}%
\thanks{$^{4}$\url{https://github.com/schmidt-ju/argoverse-navmap}}%
}

\begin{document}

\maketitle
\thispagestyle{empty}
\pagestyle{empty}

\begin{abstract}
The prediction of surrounding agents' motion is a key for safe autonomous driving.
In this paper, we explore navigation maps as an alternative to the predominant High Definition (HD) maps for learning-based motion prediction.
Navigation maps provide topological and geometrical information on road-level, HD maps additionally have centimeter-accurate lane-level information.
As a result, HD maps are costly and time-consuming to obtain, while navigation maps with near-global coverage are freely available.
We describe an approach to integrate navigation maps into learning-based motion prediction models.
To exploit locally available HD maps during training, we additionally propose a model-agnostic method for knowledge distillation.
In experiments on the publicly available Argoverse dataset with navigation maps obtained from OpenStreetMap, our approach shows a significant improvement over not using a map at all.
Combined with our method for knowledge distillation, we achieve results that are close to the original HD map-reliant models.
Our publicly available navigation map API for Argoverse enables researchers to develop and evaluate their own approaches using navigation maps$^4$.
\end{abstract}

\section{Introduction}
\copyrightnotice 
Safe autonomous driving in complex traffic scenarios requires to predict the future motion of surrounding traffic participants.
These predictions are crucial for tactical decision-making and subsequent motion planning.

Current state-of-the-art approaches for learning-based motion prediction (e.g.,~\cite{Gao2020, Liang2020, Zhou2022, Wang2022}) are developed for and evaluated on datasets that are accompanied by High Definition (HD) maps.
The core feature of HD maps is their centimeter-accurate information of lane geometry.
Obtaining and maintaining HD maps is only possible in a partially automated manner, making HD map creation costly and time-consuming~\cite{Mi2021, He2022}.
Fully automated creation fails to work in urban scenarios with complex road topologies~\cite{Joshi2015, Homayounfar2019}, scenarios with occlusion~\cite{Zhou2021} and in scenarios where important features, such as the lane markings, are hardly visible~\cite{He2022, Mattyus2016}.
In these cases, manual creation is required.
Therefore, it is questionable (i) whether global coverage can ever be achieved with such maps and (ii) whether maintaining HD maps in fast-changing environments is even possible.
Especially in urban environments, the routing of lanes can change on a daily basis, e.g., due to accidents, construction sites or road works.

\begin{figure}[thpb]
	\vspace{0.18cm} 
	\centering
	\includegraphics[width=1\columnwidth,trim={0 0 0 0.0cm},clip]{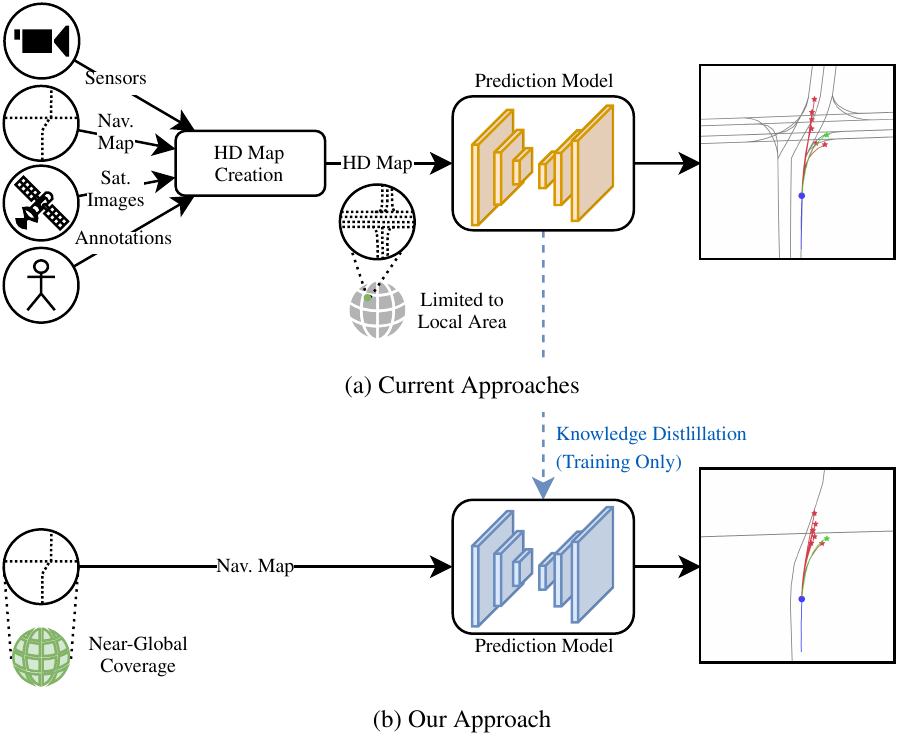}
	\caption{Comparison of (a) current approaches for motion prediction and (b) our approach. We limit the map input of motion prediction models to navigation maps with near-global coverage. Optionally, if there is an HD map during training available, we propose a teacher-student-based method for knowledge distillation. This method encourages the navigation map-aware model to extract similar embeddings from the navigation map as the teacher from the HD map.}
	\label{fig:motivation}
\end{figure}

With this paper, we want to give an incentive for a new research direction: Motion prediction with navigation maps instead of HD maps.
Navigation maps are an abstraction of HD maps.
They do not contain precise geometric information on a lane-level and are therefore more robust with regard to fast-changing environments.
Their geometric information is limited to a road-level, meaning multiple lanes are possibly summarized as one road, with varying quality.
However, they have the advantage of being widely available around the whole globe.
An example of a free navigation map with near-global coverage is the crowdsourced OpenStreetMap (OSM)~\cite{OpenStreetMap}.
To the best of our knowledge, it has never been investigated whether these navigation maps offer a viable alternative to HD maps for motion prediction.

In this work, we therefore investigate the use of navigation maps for the task of motion prediction.
We focus on road-level geometry and omit geometric information of lanes, which form the core of every HD map and serve as the main source of information in the field of prediction.
Our approach is shown in Fig.~\ref{fig:motivation}.
On the basis of two state-of-the-art trajectory prediction models, we show that using navigation maps significantly improves prediction performance over not using a map.
In the event that an HD map is locally available for the training dataset, we propose a model-agnostic teacher-student method that uses knowledge distillation to guide the training process, helping to extract relevant information from the navigation map.

In summary, our main contributions are:
\begin{itemize}
	\item We propose an approach to integrate free navigation maps into existing learning-based motion prediction models.
	\item We propose a model-agnostic teacher-student method that uses knowledge distillation to exploit locally available HD maps during training. This further improves prediction performance with navigation maps.
	\item We extensively evaluate the resulting models and our method for knowledge distillation on the publicly available Argoverse Motion Forecasting Dataset~\cite{Chang2019} and prove a performance close to their HD map-reliant counterparts.
\end{itemize}

\section{Related Work}
This section reviews related work regarding motion prediction with HD and navigation maps.
Additionally, existing work related to knowledge distillation in the field of motion prediction is described.

\subsection{Motion Prediction with High Definition Maps}
Motion prediction with HD maps is a constantly evolving topic related to autonomous driving.
There are two main groups of approaches for extracting lane information from an HD map in a learning-based manner.

Grid-based approaches~\cite{Djuric2020, Strohbeck2020, PhanMinh2020, Chai2020, Gilles2021, Kamenev2022} rasterize the information of the map, including lanes, into a multilayer Bird's-Eye View (BEV) grid.
In order to maintain the direction of lanes, the heading is most commonly color coded using the HSV color space.
Convolutional neural network backbones, such as ResNet~\cite{PhanMinh2020,Chai2020} or MobileNetV2~\cite{Strohbeck2020, Cui2019}, are then used to extract information from this grid.
The main drawback of grid-based approaches is their dependency on rasterization, which always comes with a loss of information~\cite{Liang2020}.

Vectorization-based approaches do not rely on rasterization.
A distinction between two groups of vectorization-based approaches can be made.
The first group operates on lane segments.
Lane segments are coherent sections of lanes.
Different encoders, such as graph neural networks~\cite{Gao2020, Wang2022} or a combination of 1D convolutional and recurrent neural networks~\cite{Gilles2022}, are used to form embeddings of each lane segment's geometry.
These embeddings are then included during the subsequent decoding process to obtain motion predictions.
The second group dispenses the forming of lane segment embeddings by subsampling the centerline of each lane segment to a vector-level~\cite{Liang2020, Zhou2022, Zeng2021}.
Embeddings of these vectors are then again included during the subsequent decoding process.

Motion prediction benchmarks show that vectorization-based approaches typically outperform grid-based approaches by achieving lower prediction errors~\cite{Chang2019, Ettinger2021}.

\subsection{Motion Prediction with Navigation Maps}
To the best of our knowledge, there are no publications in the field of autonomous driving that extensively evaluate motion prediction with navigation maps and compare it to HD map-reliant approaches.
CRAT-Pred~\cite{Schmidt2022} is a model specifically developed for trajectory prediction if there is no map available at all.
Using freely accessible navigation maps, which is the main idea of our work, is located between map-free approaches and approaches that rely on HD maps.

In the field of behavior planning, there are approaches that include navigation maps.
Ort et al.~\cite{Ort2018, Ort2020} describe a system to plan the future motion of the autonomous vehicle, given a navigation map from OSM and LiDAR scans.
The approach mainly relies on handcraftet rules and neglects the influence of surrounding dynamic objects, such as pedestrians and vehicles.
Xu et al.~\cite{Xu2022} describe a transformer-based approach to plan the future motion of the autonomous vehicle.
Although they refer to this as trajectory prediction, we believe that their approach is an imitation learning-based planner.
Input to the transformer is a BEV representation of the autonomous vehicle's LiDAR scan and the BEV representation of the route to follow.
They annotate the route to follow by hand, but claim that this can also be extracted from navigation maps, such as OSM.

Our approach is not intended to replace existing, sophisticated behavior planning algorithms, but rather to provide high-quality motion predictions of surrounding agents to these algorithms.

\subsection{Knowledge Distillation for Motion Prediction}
Knowledge distillation is the process of transferring the knowledge of one or multiple teacher models (teacher) into a student model (student)~\cite{Bucilua2006, Hinton2015}.
In its original use case, large unwieldy teachers are used for training a small student, achieving comparable or even better performance.
However, there are also approaches studying knowledge distillation not from the perspective of model compression, but for improving model performance~\cite{Furlanello2018}.
For this use-case, teachers and students share the same architecture.

There is few related work regarding knowledge distillation for motion prediction.
Neitz et al.~\cite{Neitz2021} apply principles of knowledge distillation to the field of motion prediction, allowing to combine advantages of model-based and model-free prediction techniques.
Monti et al.~\cite{Monti2022} use a teacher and student of the same architecture to obtain a model that is able to predict human motion with a limited amount of observed timesteps.
A higher amount of input observations are used for the teacher than for the student.
The underlying idea of using superior input representations for the teachers has its origin in the field of video processing~\cite{Bhardwaj2019, Porrello2020}.

Our approach shares this idea but applies it to a new problem.
We exploit an HD map-aware teacher in order to distill knowledge to an architecturally identical student that is only provided a navigation map.

\section{Approach} \label{sec:approach}
This section describes our approach of obtaining navigation maps and integrating them into learning-based motion prediction models.
To exploit HD maps during training, our method of knowledge distillation is used.

\subsection{Obtaining Navigation Maps}
We obtain navigation maps from OSM~\cite{OpenStreetMap}, which is a crowdsourced open geographic database.
A navigation map forms a directed graph $\mathcal{G} = \{ \mathcal{V}, \mathcal{E} \}$ consisting of nodes $v_i \in \mathcal{V}$ and edges $e_{j, i} = (v_j, v_i) \in \mathcal{E}$.
The feature vector $\mathbf{v}_i = (\mathrm{lat}_i, \mathrm{lon}_i)$ of node $v_i$ consists of its latitude and longitude in global spherical WGS84 coordinates.
Edge $e_{j, i}$ represents a road segment from node $v_j$ to node $v_i$.
We limit our navigation maps to road segments that have one of the following car-accessible types: \textit{motorway}, \textit{trunk}, \textit{primary}, \textit{secondary}, \textit{tertiary}, \textit{unclassified}, \textit{residential}, \textit{motorway\_link}, \textit{trunk\_link}, \textit{primary\_link}, \textit{secondary\_link}, \textit{tertiary\_link}, \textit{living\_street}.

Due to OSM being crowdsourced, the available features can vary for each road segment.
To show that prediction models benefit from navigation maps even if there are no road segment features available, we do not include any features for road segments in our experiments.
Adding such features is part of future work.
The geometric information of each road segment $e_{j,i}$ is given by the coordinates of its source $v_j$ and destination node $v_i$.

\subsection{Learning from Navigation Maps} \label{subsec:LearningFromNavigationMaps}
The graph structure of our navigation map representation allows for straightforward adaptation of already existing HD map-reliant prediction models.
To integrate a navigation map into such a model, the following steps are required:
\begin{itemize}
	\item Transform node feature vectors $\mathbf{v}_i$ (spherical coordinates) of the global navigation map to the local coordinate system the prediction model operates on (most commonly Cartesian coordinates), resulting in $\tilde{\mathbf{v}}_i = (x_i, y_i)$.
	\item Replace queries for lane segments in the HD map with queries for road segments (road segments are represented by edges in the graph $\mathcal{G}$) in the navigation map.
	\item Replace queries for succeeding and preceding lane segments with queries for succeeding and preceding road segments.
\end{itemize}
Our API, written for Argoverse and used in our experiments, offers these functionalities and mimics the original API for HD maps, allowing to include navigation maps with ease.

\begin{figure*}[thpb]
	\vspace{0.2cm}
	\centering
	\includegraphics[width=\textwidth, trim=0cm 0cm 1.2cm 0cm, clip]{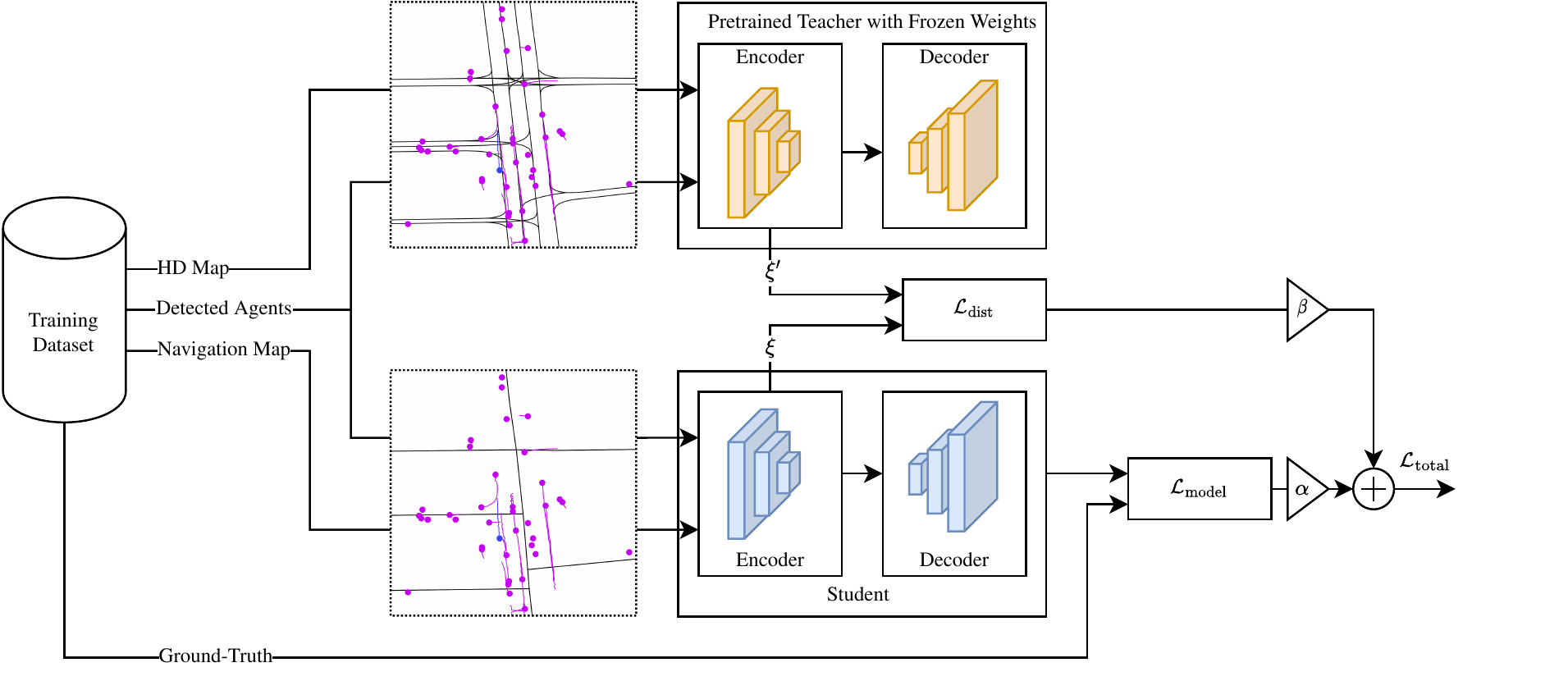}
	\caption{Overview of our model-agnostic method of using knowledge distillation to exploit HD maps during training: By means of the distillation loss $\mathcal{L}_{\mathrm{dist}}$, the latent embeddings $\xi'$ of the HD map-aware teacher model are used to guide the latent embeddings $\xi$ of the navigation map-aware student model.}
	\label{fig:teacher_forcing}
\end{figure*}

\subsection{Knowledge Distillation with HD Map-Aware Teacher} \label{subsec:TeacherStudent}
In some cases, HD maps might be available during training but unavailable during inference.
This applies to many real-world use-cases, as the recordings of datasets are often limited to a small selection of regional areas, but the prediction models trained with them are to be applied in areas that go beyond this small selection.
Obtaining HD maps for the small areas of the recordings is usually feasible, but obtaining and maintaining them with a similar level of quality globally is way more challenging.

We therefore propose a model-agnostic knowledge distillation method to improve prediction performance with navigation maps if HD maps are available during training.
Model-agnostic means that this method can be applied to any already existing and upcoming learning-based prediction model.

Our method consists of a teacher and a student prediction model.
An overview is given in Fig.~\ref{fig:teacher_forcing}.
Teacher and student share the same architecture but are provided with different input information.
The teacher is provided with information of the HD map and the student is provided with information of the navigation map.

Every learning-based prediction model that relies on map information has a stage where information of the agent to be predicted and the map are fused together into one latent embedding $\xi$.
Ways to perform this fusion step include attention (e.g.,~\cite{Liang2020, Zhou2022}) or concatenation (e.g.,~\cite{Djuric2020, Strohbeck2020, Gilles2021}).
Our method exploits this property of prediction models and uses the latent embedding $\xi' \in \mathcal{R}^{d_{\mathrm{t}}}$ generated by the fully trained teacher to guide the latent embedding $\xi \in \mathcal{R}^d, d \geq {d_{\mathrm{t}}}$ of the student during training.
To obtain this behavior, we define the distillation loss as
\begin{equation}
	\mathcal{L}_{\mathrm{dist}}(\xi', \xi) = \frac{1}{{d_{\mathrm{t}}}} \sum_{i=0}^{{d_{\mathrm{t}}}-1}({\xi'}_i - {\xi}_i)^2 \text{.}
\end{equation}

During the training of the student, the total loss is then calculated as
\begin{equation}
\mathcal{L}_{\mathrm{total}} = \alpha \cdot \mathcal{L}_{\mathrm{model}} + \beta \cdot \mathcal{L}_{\mathrm{dist}} \text{,}
\end{equation}
with $\mathcal{L}_{\mathrm{model}}$ referring to the original loss of the prediction model.
$\alpha$ and $\beta$ are scaling factors.

By introducing the distillation loss, the student is encouraged to form map-aware embeddings $\xi$ that are similar to the one obtained by the teacher that has access to the HD map.
Accordingly, the exploitation of the HD map during training is carried out directly in the latent space and requires no handcrafted preparation of the input data.

\section{Experiments}
This section describes our evaluation on the publicly available Argoverse Motion Forecasting Dataset~\cite{Chang2019}, subsequently referred to as Argoverse.
We evaluate our OSM-based approach with two state-of-the-art trajectory prediction models.

\subsection{Dataset}
In contrast to many other datasets, including the recent Argoverse 2 dataset~\cite{Wilson2021}, the Argoverse dataset contains information about the origins of its local Cartesian coordinate systems.
As described in this section, this allows to transform the spherical coordinates of navigation maps into the local coordinate system used by Argoverse.

\subsubsection{Dataset Properties}
Argoverse is a trajectory prediction dataset consisting of sequences recorded in Miami and Pittsburgh.
Each sequence has a duration of five seconds, is sampled with \SI{10}{Hz} and contains the position of tracked vehicles in the surroundings of an autonomous vehicle.
$201$k sequences form the training split and $39$k form the validation split.
Given the first two seconds of a sequence, the goal is to predict the trajectory of the subsequent three seconds of one predefined target vehicle.
Two HD maps, one covering relevant areas of Miami and one covering relevant areas of Pittsburgh, are provided in the dataset.

\subsubsection{Coordinate System}
All coordinates are given in local Cartesian city coordinates.
The following transformations are applied to transform global spherical coordinates $\mathbf{v}_i = (\mathrm{lat}_i, \mathrm{lon}_i)$ of a navigation map $\mathcal{G}$ into local city coordinates $\tilde{\mathbf{v}}_i = (x_i, y_i)$:
\begin{itemize}
	\item Transform global spherical coordinates into Universal Transverse Mercator (UTM) coordinates.
	\item Substract the origin of the local city coordinate system of the resulting UTM coordinates.
\end{itemize}

Origins of the local city coordinate systems of Miami and Pittsburgh are given in the original publication~\cite{Chang2019}.\footnote{Miami: $580560.0088$ Easting, $2850959.9999$ Northing, Zone $17$; \\ Pittsburgh: $583710.0070$ Easting, $4477259.9999$ Northing, Zone $17$}

\subsection{Metrics}
We follow the standard protocol for the evaluation of trajectory prediction models and use the minimum Average Displacement Error (minADE), minimum Final Displacement Error (minFDE) and Miss Rate (MR) for single mode ($k=1$) and multi-modal predictions ($k=6$).

minFDE is the lowest Euclidean distance of the $k$ predicted endpoints and the ground-truth endpoint.
minADE is the average Euclidean distance of the trajectory selected for the minFDE calculation and the ground-truth trajectory.
MR is the ratio of sequences where none of the $k$ predicted endpoints is within a radius of \SI{2}{m} to the ground-truth endpoint.
Metrics are averaged over the corresponding dataset split.

\subsection{Model Implementation Details}
We evaluate our approach using two state-of-the-art trajectory prediction models with publicly available code.
Model architectures and training protocols are the same as in the original publications.
To obtain the geometry of a road segment $e_{j, i}$, we linearly interpolate between $\tilde{\mathbf{v}}_j$ and $\tilde{\mathbf{v}}_i$ with a step size of \SI{2}{m}.

We evaluate two different variants of our teacher-student method per prediction model.
The first variant uses $d = {d_{\mathrm{t}}}$, meaning that the student is encouraged to approximate the full HD map-aware embeddings $\xi'$ of the teacher.
The second variant uses $d =1.5\cdot {d_{\mathrm{t}}}$.
This results in a shared embedding: ${d_\mathrm{t}}$ features are used to approximate $\xi'$ and ${d_{\mathrm{t}}} \cdot 0.5$ features are unguided.
Using additional unguided features allows the student to also extract navigation map-specific information.
All experiments are performed with $\alpha = \beta = 1$.

\input{tables/results_lanegcn}
\input{tables/results_hivt}
\begin{figure}[thpb]
	\includegraphics[width=1\columnwidth,trim={0 0 0 0.0cm},clip]{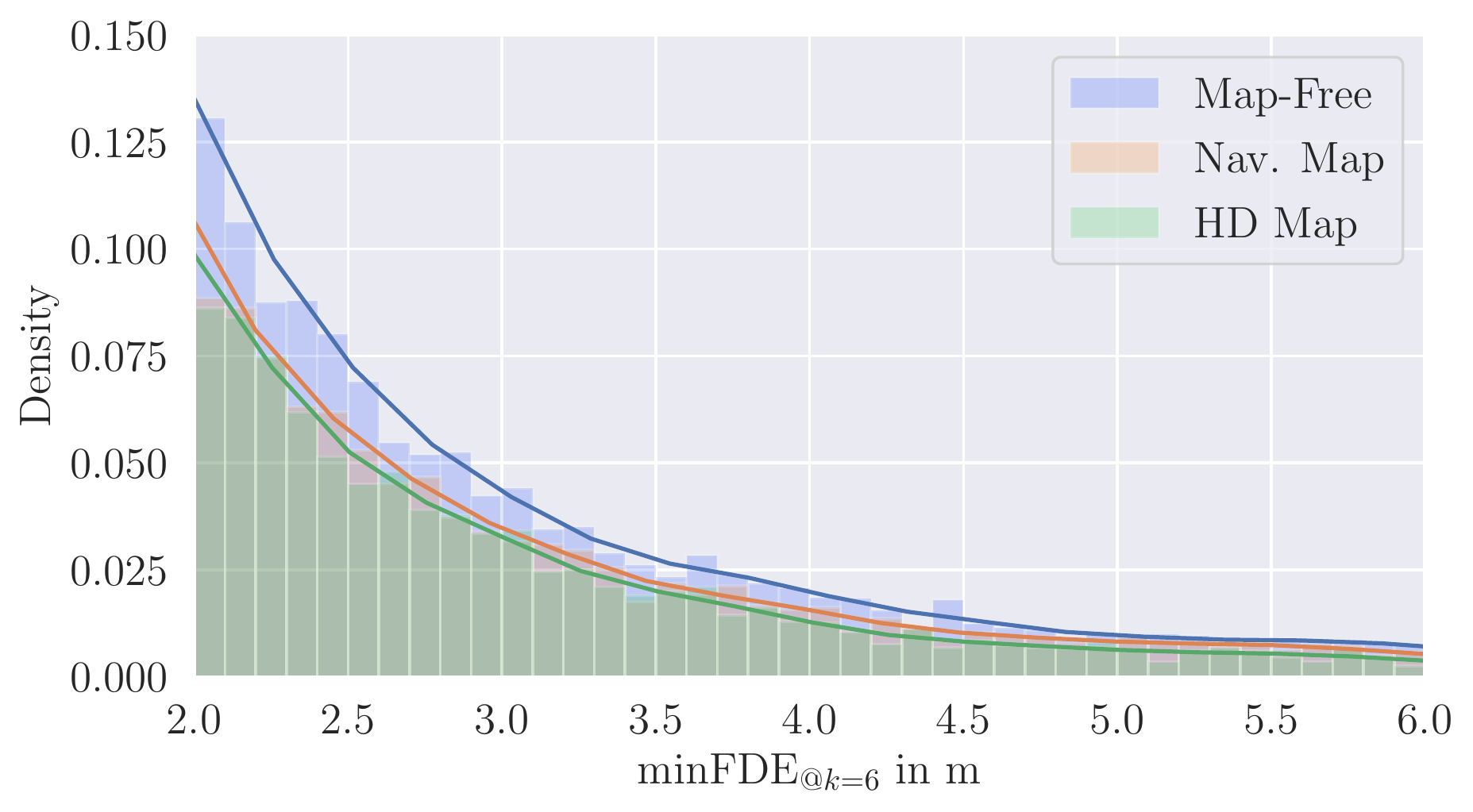}
	\caption{Normalized histogram of the minFDE$_{@k=6}$ of different LaneGCN variants.}
	\label{fig:fde_histogram}
\end{figure}

\subsubsection{LaneGCN}
LaneGCN~\cite{Liang2020} originally uses graph-based encoding of lane geometry.
Four attention-based fusion mechanisms, namely Actor-Lane, Lane-Lane, Lane-Actor and Actor-Actor, aggregate information into one latent embedding per agent in a scene.
Subsequently, linear residual layers are used to predict multiple trajectories with corresponding confidences.

We use our developed Argoverse API for navigation maps and adapt LaneGCN with our approach described in Section~\ref{subsec:LearningFromNavigationMaps}.
This results in the graph-based encoding of lane geometry being replaced by a graph-based encoding of road geometry.
For our teacher-student method described in Section~\ref{subsec:TeacherStudent}, we use the agent-wise embeddings $\xi$ resulting from the Lane-Actor fusion.
In LaneGCN, this corresponds to the stage where information of the agent to be predicted and the map are fused together.

\subsubsection{HiVT}
HiVT~\cite{Zhou2022} originally uses cross-attention to fuse information of agents with geometric information of surrounding lanes.
Predictions and confidences are generated by multilayer perceptrons.

Again, we use our developed Argoverse API for navigation maps and adapt HiVT with our proposed approach, resulting in geometry on a lane-level being replaced by geometry on a road-level.
For our teacher-student method, the agent-wise embeddings $\xi$ resulting from the fusion of agent information with map information via cross-attention are used.

\begin{figure*}[thpb]
	\vspace{-0.1cm}
	\centering
	\subfloat{%
		\begin{overpic}[width=0.33\textwidth, trim=6cm 10cm 15cm 10cm, clip, frame]
			{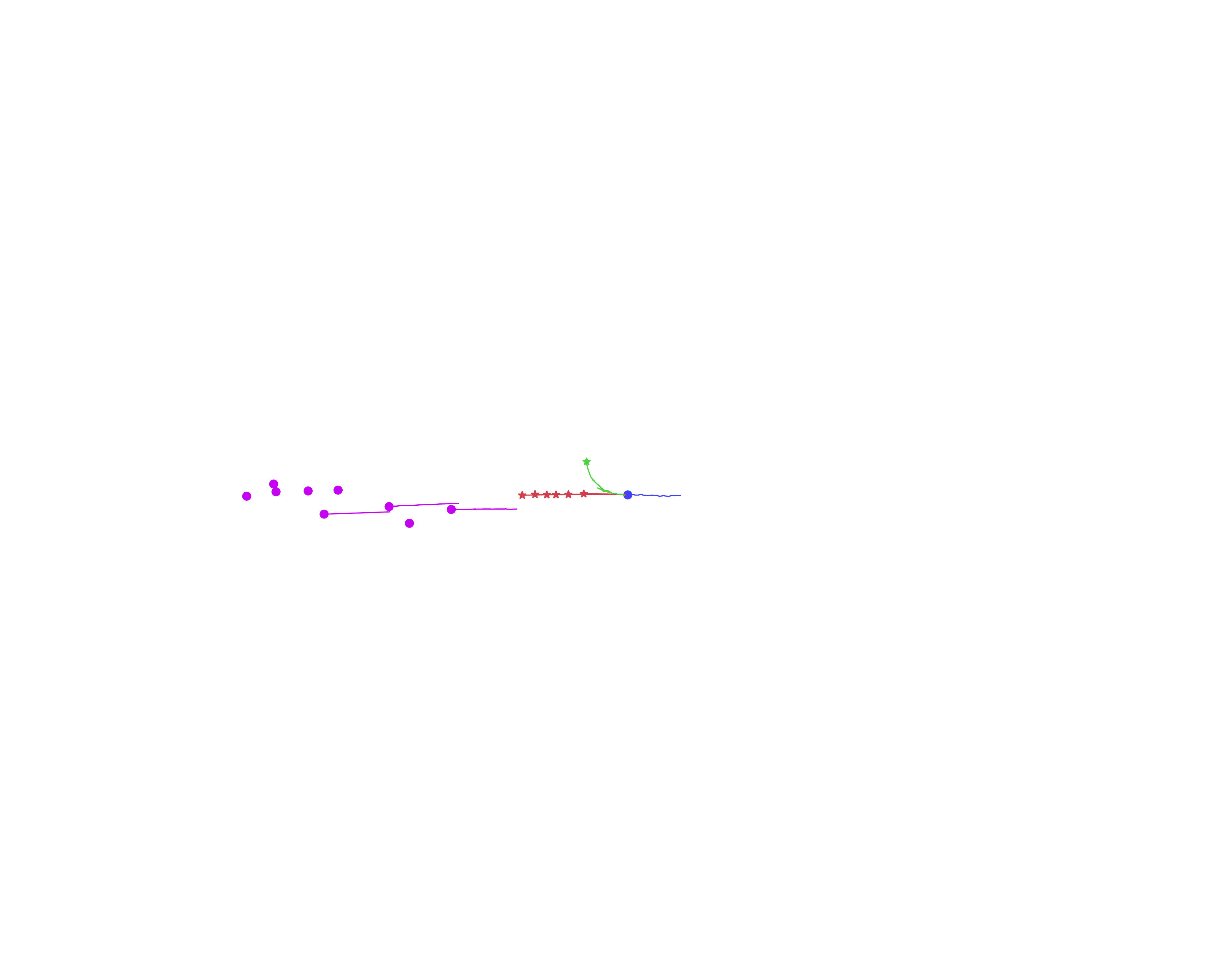}
			\setlength\fboxsep{0pt}
			\footnotesize
			\put(1, 52){\pgfsetfillopacity{0.8}\colorbox{white}{\minibox[frame]{\pgfsetfillopacity{1}Seq. 1 \\ Map-Free}}}
	\end{overpic}}
	\hfill
	\subfloat{%
		\begin{overpic}[width=0.33\textwidth, trim=6cm 10cm 15cm 10cm, clip, frame]
			{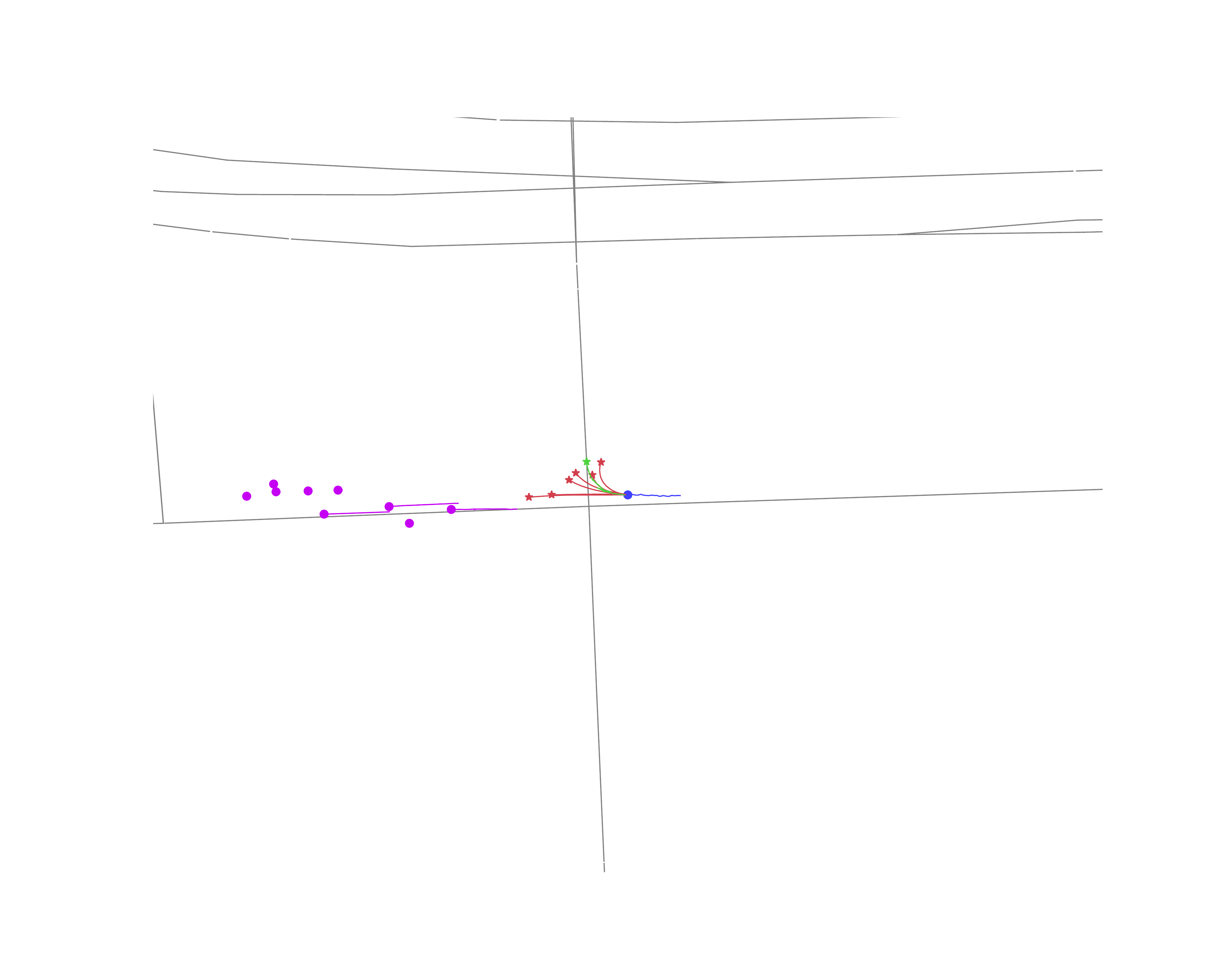}
			\setlength\fboxsep{0pt}
			\footnotesize
			\put(1, 52){\pgfsetfillopacity{0.8}\colorbox{white}{\minibox[frame]{\pgfsetfillopacity{1}Seq. 1 \\ Nav. Map}}}
	\end{overpic}}
	\hfill
	\centering
	\subfloat{%
		\begin{overpic}[width=0.33\textwidth, trim=6cm 10cm 15cm 10cm, clip, frame]
			{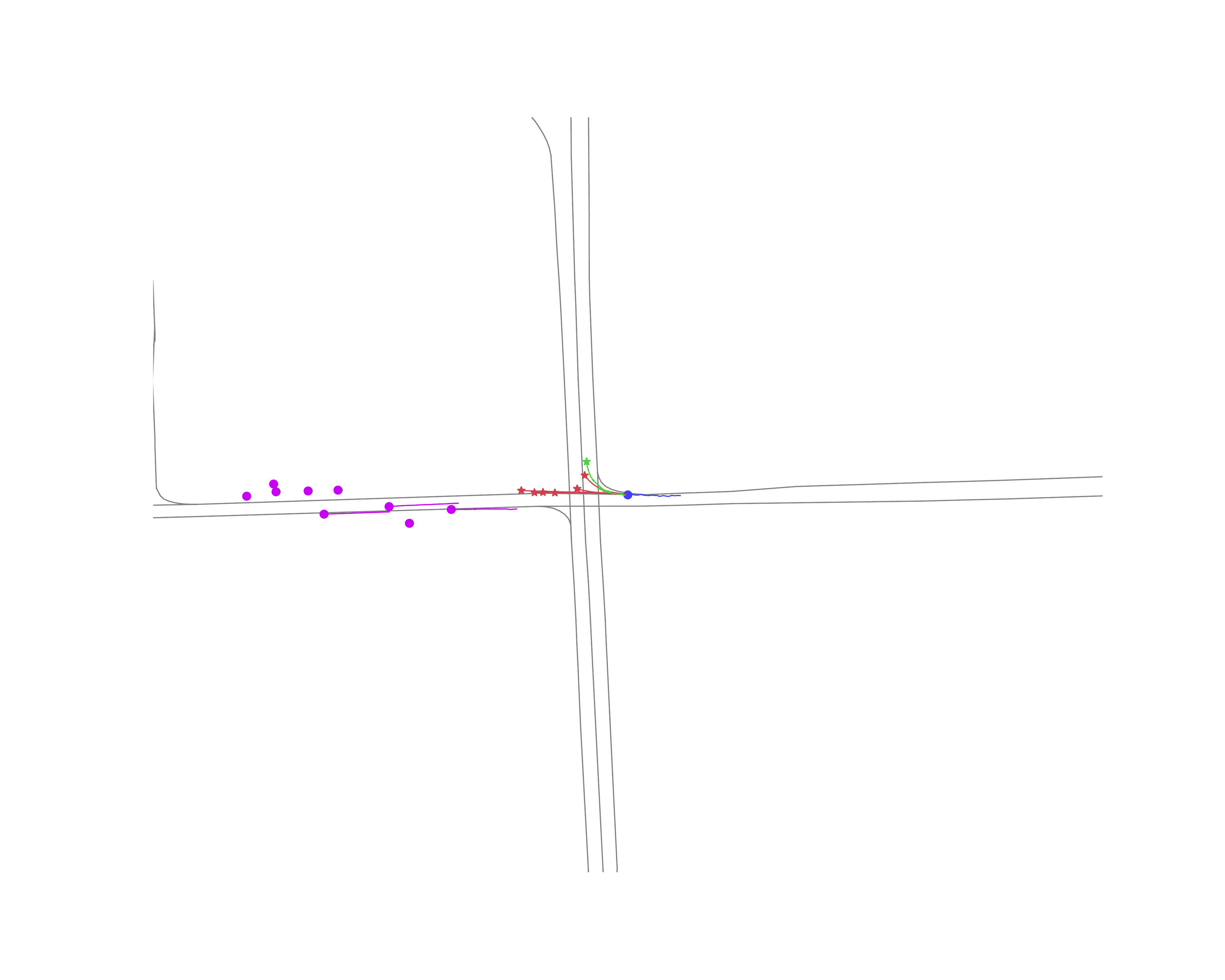}
			\setlength\fboxsep{0pt}
			\footnotesize
			\put(1, 52){\pgfsetfillopacity{0.8}\colorbox{white}{\minibox[frame]{\pgfsetfillopacity{1}Seq. 1 \\ HD Map}}}
	\end{overpic}}
	\\[-0.26cm]
	\subfloat{%
		\begin{overpic}[width=0.33\textwidth, trim=13cm 10cm 8cm 10cm, clip, frame]
			{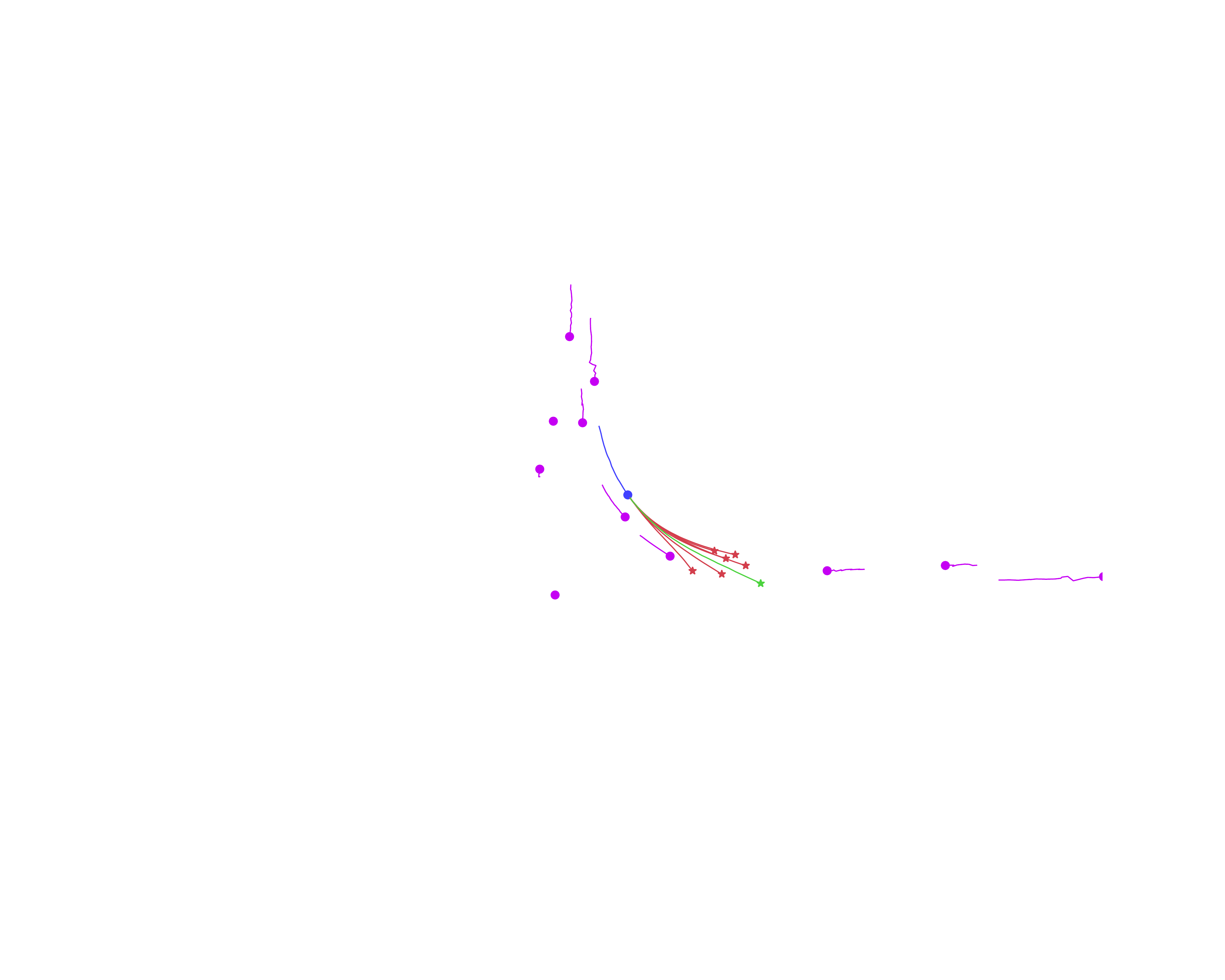}
			\setlength\fboxsep{0pt}
			\footnotesize
			\put(1, 52){\pgfsetfillopacity{0.8}\colorbox{white}{\minibox[frame]{\pgfsetfillopacity{1}Seq. 2 \\ Map-Free}}}
	\end{overpic}}
	\hfill
	\subfloat{%
		\begin{overpic}[width=0.33\textwidth, trim=13cm 10cm 8cm 10cm, clip, frame]
			{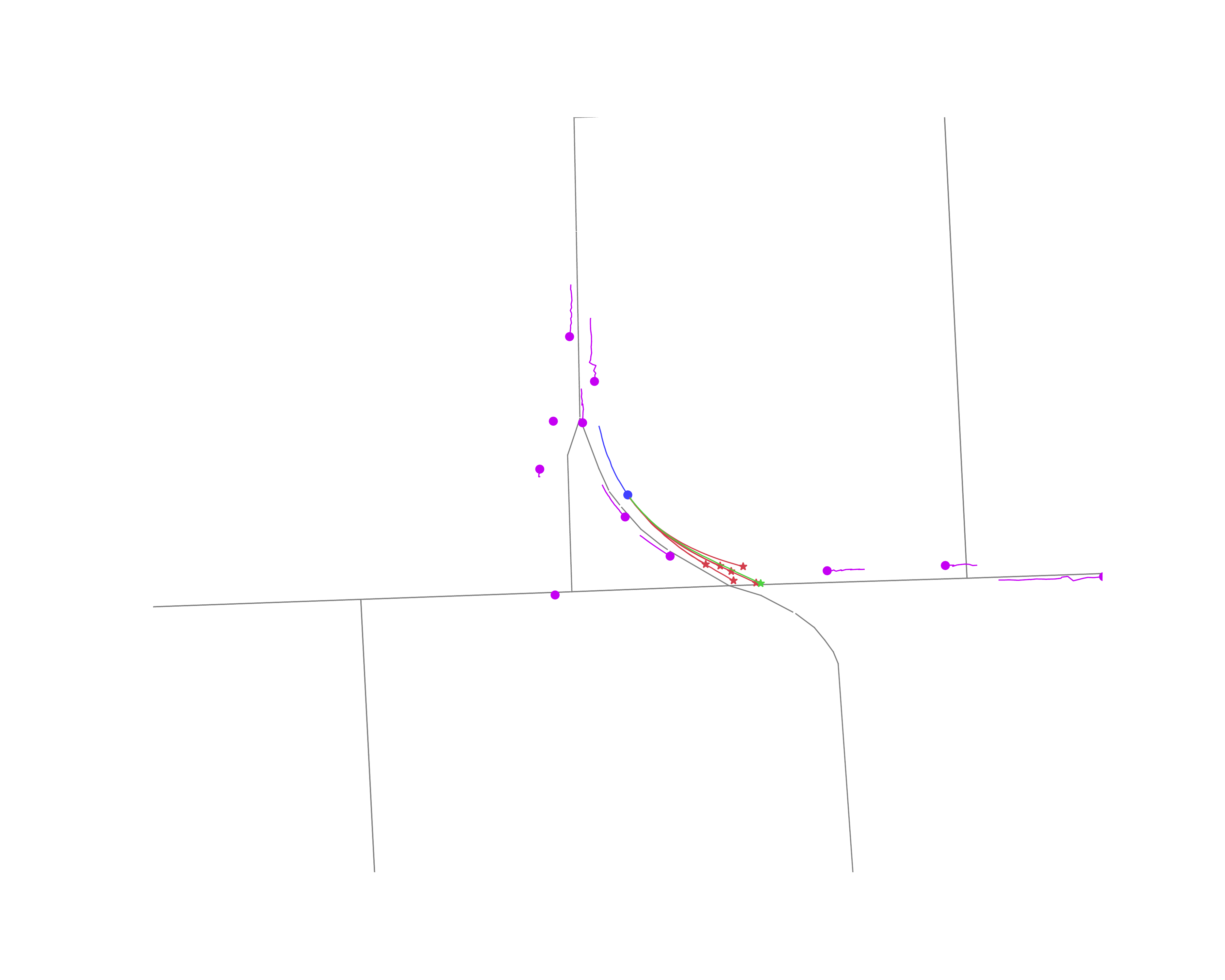}
			\setlength\fboxsep{0pt}
			\footnotesize
			\put(1, 52){\pgfsetfillopacity{0.8}\colorbox{white}{\minibox[frame]{\pgfsetfillopacity{1}Seq. 2 \\ Nav. Map}}}
	\end{overpic}}
	\hfill
	\centering
	\subfloat{%
		\begin{overpic}[width=0.33\textwidth, trim=13cm 10cm 8cm 10cm, clip, frame]
			{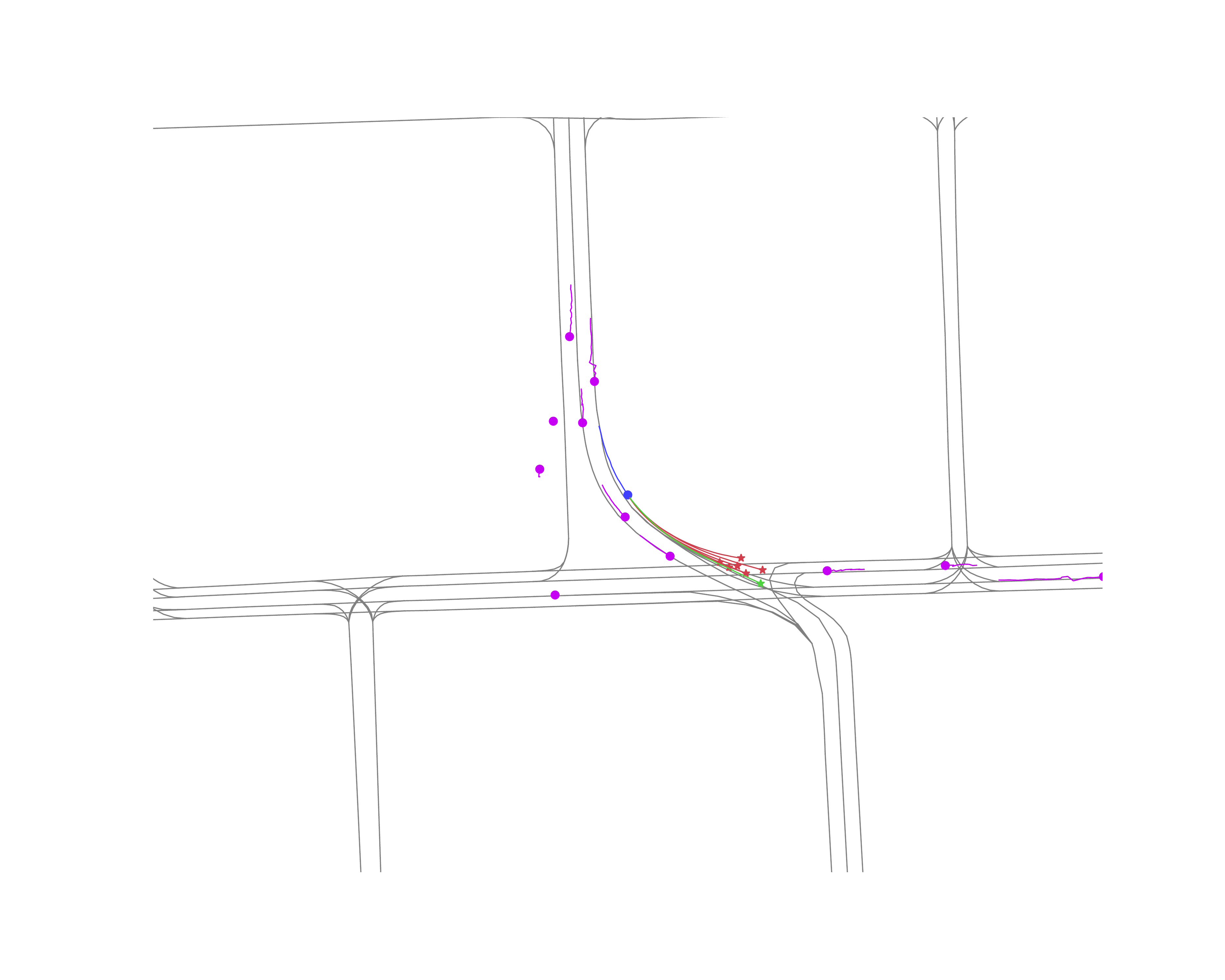}
			\setlength\fboxsep{0pt}
			\footnotesize
			\put(1, 52){\pgfsetfillopacity{0.8}\colorbox{white}{\minibox[frame]{\pgfsetfillopacity{1}Seq. 2 \\ HD Map}}}
	\end{overpic}}
	\\[-0.26cm]
	\subfloat{%
		\begin{overpic}[width=0.33\textwidth, trim=13cm 10cm 8cm 10cm, clip, frame]
			{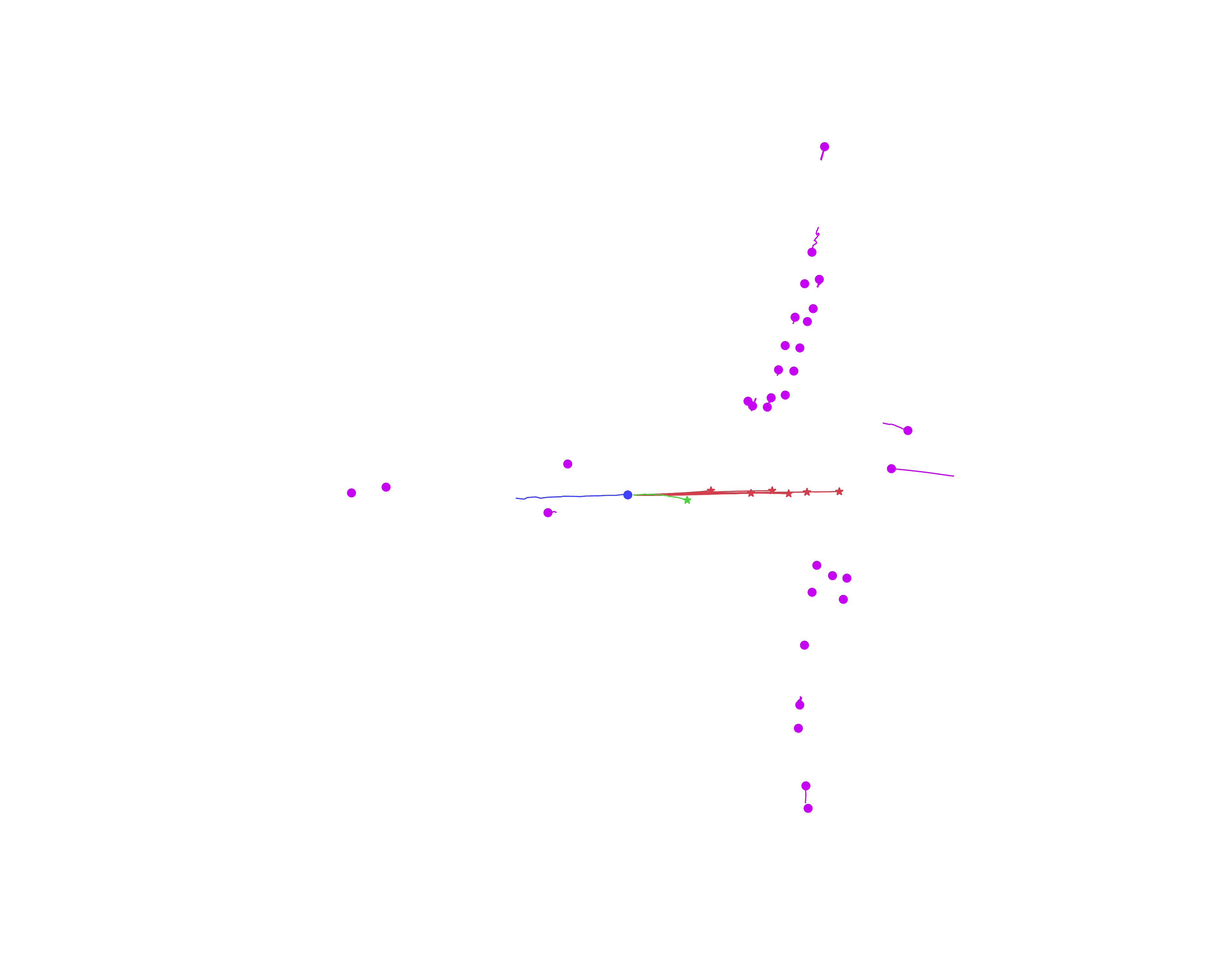}
			\setlength\fboxsep{0pt}
			\footnotesize
			\put(1, 52){\pgfsetfillopacity{0.8}\colorbox{white}{\minibox[frame]{\pgfsetfillopacity{1}Seq. 3 \\ Map-Free}}}
	\end{overpic}}
	\hfill
	\subfloat{%
		\begin{overpic}[width=0.33\textwidth, trim=13cm 10cm 8cm 10cm, clip, frame]
			{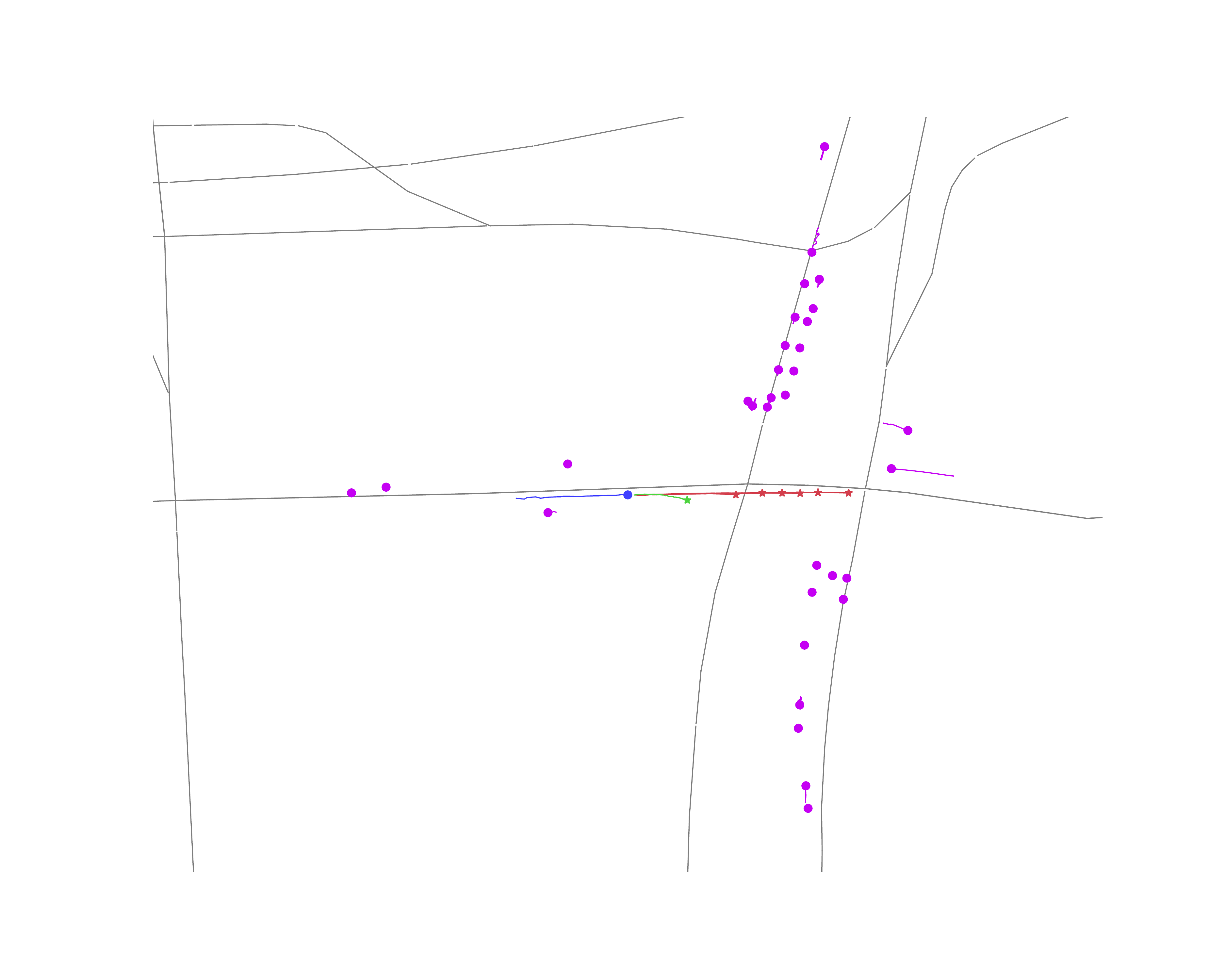}
			\setlength\fboxsep{0pt}
			\footnotesize
			\put(1, 52){\pgfsetfillopacity{0.8}\colorbox{white}{\minibox[frame]{\pgfsetfillopacity{1}Seq. 3 \\ Nav. Map}}}
	\end{overpic}}
	\hfill
	\centering
	\subfloat{%
		\begin{overpic}[width=0.33\textwidth, trim=13cm 10cm 8cm 10cm, clip, frame]
			{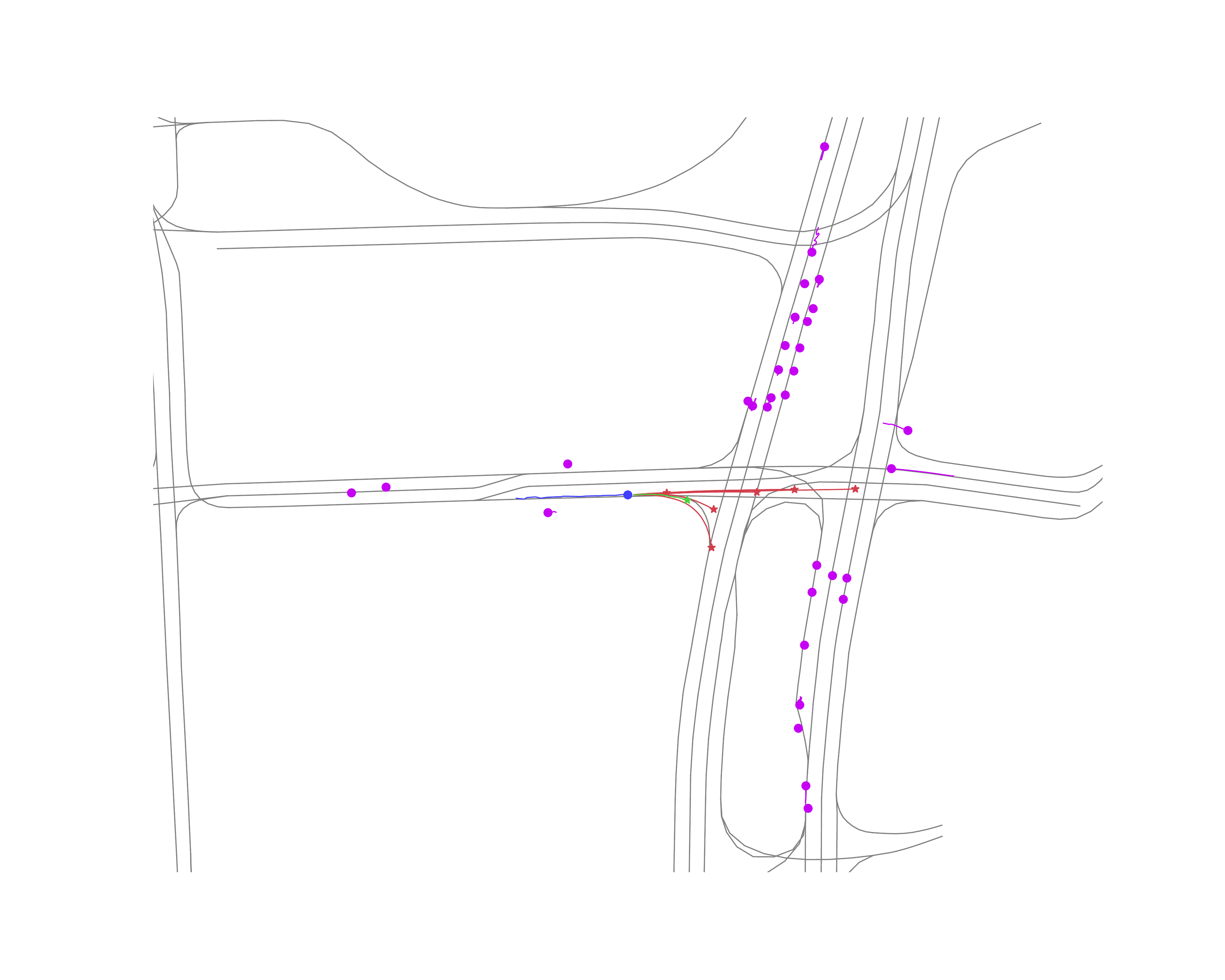}
			\setlength\fboxsep{0pt}
			\footnotesize
			\put(1, 52){\pgfsetfillopacity{0.8}\colorbox{white}{\minibox[frame]{\pgfsetfillopacity{1}Seq. 3 \\ HD Map}}}
	\end{overpic}}
	\caption{Exemplary sequences of LaneGCN on the Argoverse validation set, trained and validated map-free (left), with our navigation map-aware teacher-student method $d =1.5\cdot d_\mathrm{t}$ (center) and with the HD map (right). Three sequences are shown. The past observed trajectory of the vehicle to be predicted is colored in blue, the ground-truth future trajectory in green. Predictions are colored in red. Past trajectories of other vehicles are colored in purple. The map-information that is used by the corresponding model is colored in gray.}
	\label{fig:qualitative_results}
\end{figure*}

\subsection{Quantitative Results}
Table~\ref{tab:results_lanegcn} and Table~\ref{tab:results_hivt} show the quantitative results of our approach applied to LaneGCN and HiVT on the Argoverse validation set.
For both models, the results of our approach are compared with the original models trained map-free and with HD map.

The results of LaneGCN and HiVT both show that integrating navigation maps significantly improves performance compared to the map-free models.
Our proposed model-agnostic teacher-student method that uses knowledge distillation to exploit HD maps during training improves performance even more.
Especially for LaneGCN, the $d=1.5\cdot d_\mathrm{t}$ variant outperforms all other navigation map-reliant variants.
The same statement, although with slightly less improvement, applies to HiVT.

For HiVT, it is observable that the difference between the variant without ($d=96$) and with our teacher-student method ($d=1.5\cdot d_\mathrm{t}, d_\mathrm{t}=64$) is small.
However, the performance of both is close to the original HD map-reliant variant.

In general, the results of the navigation map-reliant variants are close to the results of the original HD map-reliant variant.
For instance, the minFDE$_{@k=6}$ difference between our best navigation map-reliant LaneGCN variant and the original HD map-reliant variant for a \SI{3}{s} prediction horizon is \SI{9}{cm} on average.
For HiVT, our teacher-student method even results in an identical MR$_{@k=1}$ as the original HD map-reliant variant.

Fig.~\ref{fig:fde_histogram} shows a normalized histogram of the minFDE$_{@k=6}$ of LaneGCN beyond the Miss Rate distance of \SI{2}{m}.
Kernel density estimation is used to obtain a smoothed trend.
One distinct observation is that our navigation map-reliant variant ($d =1.5\cdot {d_{\mathrm{t}}}$) consistently leads to less sequences with high prediction errors than the map-free variant.
This suggests that the integration of navigation maps reduces sequences in which the intent of the predicted agent was not correctly predicted at all (e.g., failing to predict a turn or a lane change maneuver).

\subsection{Qualitative Results}
Our qualitative results are generated with different LaneGCN variants on the Argoverse validation set.
We reviewed over $400$ sequences and selected three interesting ones for visualization in Fig.~\ref{fig:qualitative_results}.
While there are many positive examples, we only observed two clear negative examples.
One of them is illustrated as the third sequence in the figure.
For each sequence, the results of the map-free variant (left), the adaptation with our navigation map-reliant teacher-student method (center) and the original HD map-reliant variant (right) are shown.
Color codes are given in the caption of the figure.

The first sequence shows a turn maneuver.
While the map-free variant fails to predict this turn maneuver, both map-reliant variants are able to correctly predict both possible intents of the vehicle.
The second sequence is another positive example.
Both map-reliant variants are able to precisely predict the movement along the curved road.
Due to having no information about the underlying road or lane topology and geometry, the predictions of the map-free variant are less precise.
The third sequence illustrates a negative example.
Both, the map-free and the navigation map-reliant variant, fail to predict the turn maneuver.
One possible explanation for failure of the navigation map-reliant variant is the acute angle of the road into which the turn is made.

\section{Conclusion}
In this paper, we explore the use of navigation maps as an alternative to HD maps for motion prediction.
By implementing the proposed approach for two learning-based trajectory prediction models, we prove a significant improvement in prediction performance compared to using no map.
In combination with our model-agnostic knowledge distillation method that exploits HD maps during training, results that are close to the HD map-reliant counterparts are achieved.
The publicly available source code of our navigation map API for Argoverse enables other researchers to develop and evaluate their own navigation map-based approaches for motion prediction with ease.

It remains to be investigated whether similar results are achievable using navigation maps in other HD map-reliant application areas beyond motion prediction, for instance traffic scene reasoning~\cite{MonningerSchmidt2023}.

\addtolength{\textheight}{-0.cm}

\bibliographystyle{IEEEtran}
\bibliography{Literature}

\end{document}

%% file: tables/results_lanegcn.tex
\begin{table*}[thpb]
	\vspace{0.25cm}
	\caption{Results of our approach applied to LaneGCN}
	\label{tab:results_lanegcn}
	\setlength{\tabcolsep}{3.8pt}
	\centering
	\begin{tabularx}{1\textwidth}{Xllllllllllll}
		\toprule
		\multirow{2}{*}{Model Variant}                                   & \multicolumn{2}{c}{HD Map} & \multicolumn{2}{c}{Nav. Map} & \multicolumn{2}{c}{Embed. Size} &              \multicolumn{3}{c}{$k=1$}              &              \multicolumn{3}{c}{$k=6$}              \\
		                                                                 & Train      & Val           & Train      & Val             & $d_\mathrm{t}$ & $d$            & minADE          & minFDE          & MR              & minADE          & minFDE          & MR              \\ \midrule
		Original (Map-Free)                                              &            &               &            &                 & -              & $128$          & $1.52$          & $3.44$          & $0.54$          & $0.78$          & $1.28$          & $0.14$          \\ \midrule
		Ours (w/ Nav. Map, Small Embedding)                              &            &               & \checkmark & \checkmark      & -              & $128$          & $1.43$          & $3.17$          & $0.51$          & $0.76$          & $1.21$          & $\mathbf{0.12}$ \\
		Ours (w/ Nav. Map, Large Embedding)                              &            &               & \checkmark & \checkmark      & -              & $192$          & $1.42$          & $3.16$          & $0.51$          & $0.76$          & $1.21$          & $0.13$          \\
		Ours (w/ Nav. Map + Teacher-Student, $d = d_\mathrm{t}$)         & \checkmark &               & \checkmark & \checkmark      & $128$          & $128$          & $1.40$          & $3.12$          & $\mathbf{0.50}$ & $0.75$          & $1.20$          & $\mathbf{0.12}$ \\
		Ours (w/ Nav. Map + Teacher-Student, $d =1.5\cdot d_\mathrm{t}$) & \checkmark &               & \checkmark & \checkmark      & $128$          & $192$          & $\mathbf{1.38}$ & $\mathbf{3.07}$ & $\mathbf{0.50}$ & $\mathbf{0.74}$ & $\mathbf{1.17}$ & $\mathbf{0.12}$ \\ \midrule
		Original (w/ HD Map)                                             & \checkmark & \checkmark    &            &                 & -              & $128$          & $1.34$          & $2.96$          & $0.49$          & $0.71$          & $1.08$          & $0.10$          \\ \bottomrule
	\end{tabularx}
\end{table*}

%% file: tables/results_hivt.tex
\begin{table*}[thpb]
	\caption{Results of our approach applied to HiVT}
	\label{tab:results_hivt}
	\setlength{\tabcolsep}{3.8pt}
	\centering
	\begin{tabularx}{1\textwidth}{Xllllllllllll}
		\toprule
		\multirow{2}{*}{Model Variant}                                   & \multicolumn{2}{c}{HD Map} & \multicolumn{2}{c}{Nav. Map} & \multicolumn{2}{c}{Embed. Size} &              \multicolumn{3}{c}{$k=1$}              &              \multicolumn{3}{c}{$k=6$}              \\
		                                                                 & Train      & Val           & Train      & Val             & $d_\mathrm{t}$ & $d$            & minADE          & minFDE          & MR              & minADE          & minFDE          & MR              \\ \midrule
		Original (Map-Free)                                              &            &               &            &                 & -              & $64$           & $1.48$          & $3.27$          & $0.52$          & $0.76$          & $1.24$          & $0.14$          \\ \midrule
		Ours (w/ Nav. Map, Small Embedding)                              &            &               & \checkmark & \checkmark      & -              & $64$           & $1.43$          & $3.14$          & $0.50$          & $0.73$          & $1.15$          & $0.12$          \\
		Ours (w/ Nav. Map, Large Embedding)                              &            &               & \checkmark & \checkmark      & -              & $96$           & $\mathbf{1.41}$ & $3.09$          & $0.49$          & $\mathbf{0.71}$ & $1.11$          & $\mathbf{0.11}$ \\
		Ours (w/ Nav. Map + Teacher-Student, $d = d_\mathrm{t}$)         & \checkmark &               & \checkmark & \checkmark      & $64$           & $64$           & $1.44$          & $3.16$          & $0.49$          & $0.73$          & $1.15$          & $0.12$          \\
		Ours (w/ Nav. Map + Teacher-Student, $d =1.5\cdot d_\mathrm{t}$) & \checkmark &               & \checkmark & \checkmark      & $64$           & $96$           & $\mathbf{1.41}$ & $\mathbf{3.08}$ & $\mathbf{0.48}$ & $\mathbf{0.71}$ & $\mathbf{1.09}$ & $\mathbf{0.11}$ \\ \midrule
		Original (w/ HD Map)                                             & \checkmark & \checkmark    &            &                 & -              & $64$           & $1.36$          & $2.98$          & $0.48$          & $0.69$          & $1.03$          & $0.10$          \\ \bottomrule
	\end{tabularx}
\end{table*}